\documentclass[letterpaper]{article} 
\usepackage{aaai25}  
\usepackage{times}  
\usepackage{helvet}  
\usepackage{courier}  
\usepackage[hyphens]{url}  
\usepackage{graphicx} 
\usepackage{natbib}  
\usepackage{caption} 
\usepackage{algorithm}
\usepackage{algorithmic}
\usepackage{subcaption}
\usepackage{amsmath,amssymb}
\usepackage{newfloat}
\usepackage{listings}
\DeclareCaptionStyle{ruled}{labelfont=normalfont,labelsep=colon,strut=off} 
\floatstyle{ruled}
\newfloat{listing}{tb}{lst}{}
\floatname{listing}{Listing}
\title{Exploring Visual Embedding Spaces Induced by\\Vision Transformers for Online Auto Parts Marketplaces}
\author{
    Cameron Armijo,
    Pablo Rivas
}
\affiliations{
    Department of Computer Science\\
    Baylor University, Texas, USA\\
    \{Cameron\_Armijo1, Pablo\_Rivas\}@Baylor.edu
}

\usepackage{bibentry}

\begin{document}

\maketitle

\begin{abstract}
This study examines the capabilities of the Vision Transformer (ViT) model in generating visual embeddings for images of auto parts sourced from online marketplaces, such as Craigslist and OfferUp. By focusing exclusively on single-modality data, the analysis evaluates ViT's potential for detecting patterns indicative of illicit activities. The workflow involves extracting high-dimensional embeddings from images, applying dimensionality reduction techniques like Uniform Manifold Approximation and Projection (UMAP) to visualize the embedding space, and using K-Means clustering to categorize similar items. Representative posts nearest to each cluster centroid provide insights into the composition and characteristics of the clusters. While the results highlight the strengths of ViT in isolating visual patterns, challenges such as overlapping clusters and outliers underscore the limitations of single-modal approaches in this domain. This work contributes to understanding the role of Vision Transformers in analyzing online marketplaces and offers a foundation for future advancements in detecting fraudulent or illegal activities.
\end{abstract}

\section{Introduction}
The transformer architecture, originally developed for natural language processing (NLP), has proven highly successful in a variety of tasks, from language translation \cite{vaswani2023attentionneed} to text generation \cite{Radford2018ImprovingLU}. Its core strength lies in the self-attention mechanism, which enables the model to capture relationships between all parts of an input sequence simultaneously. Building on the success of transformers, this architecture has been adapted to computer vision through Vision Transformers. These models take the same transformer principles—such as self-attention and tokenization—and apply them to image data by dividing images into fixed-size patches that are treated as input tokens. Vision Transformers have reached state-of-the-art benchmarks in computer vision, comparable to and even exceeding CNNs \cite{dosovitskiy2021imageworth16x16words}. The adaptability of transformers to visual data demonstrates their versatility and effectiveness, opening new avenues for image analysis tasks such as object detection with models like DETR \cite{carion2020endtoendobjectdetectiontransformers, shehzadi2023objectdetectiontransformersreview} and image synthesis with models such as DCGAN \cite{radford2016unsupervisedrepresentationlearningdeep, goodfellow2014generativeadversarialnetworks}. These models are derivatives of the original Vision Transformer architecture, each tailored for their specific tasks.

\begin{figure}[b!]
    \centering
    \includegraphics[width=0.5\columnwidth]{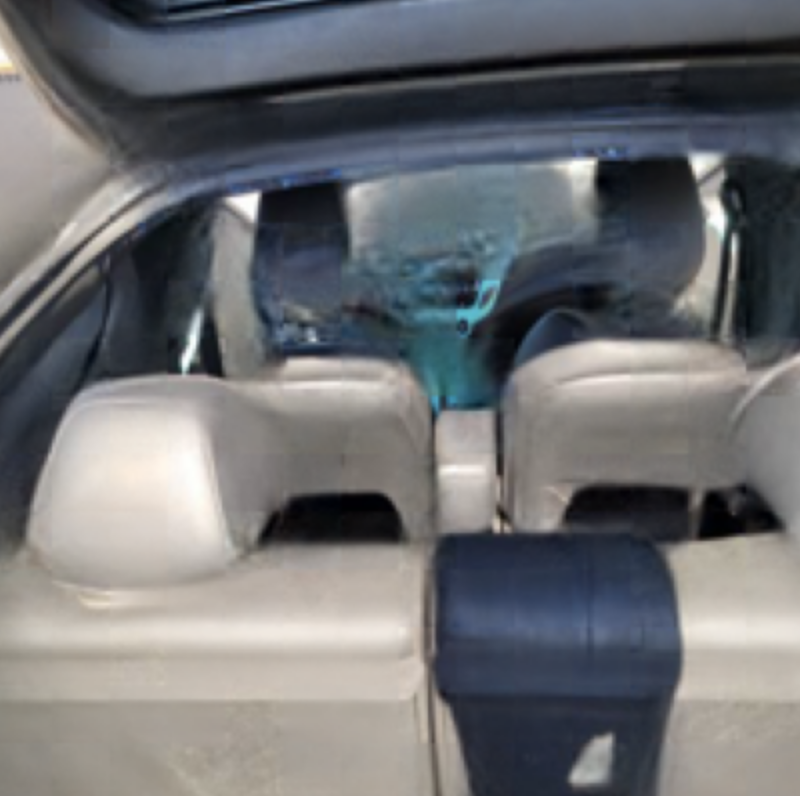}
    \caption{Example of an interior view from an online auto parts listing, showcasing a car's seating and dashboard. This image represents the type of visual data used for clustering and analysis in this study.}
    \label{postExample}
\end{figure}

This study focuses exclusively on image classification tasks using ViT-Base, a single-modal deep learning model based on the Vision Transformer architecture. Specifically, we fine-tune ViT-Base on a dataset comprising images from online auto parts listings to evaluate its computer vision capabilities. Due to the nature of these online listings, only image data is considered (single-modality), while textual data such as captions from these posts is excluded. Figure~\ref{postExample} illustrates an example image from the dataset, highlighting the typical content of these listings.

While previous works have demonstrated the superiority of multimodal approaches \cite{hamara2024latentenginemanifoldsanalyzing,rashid2024aisafetypracticeenhancing} using models like ImageBind and OpenFlamingo, this study specifically evaluates a single-modal approach. This analysis is significant in understanding the limitations and potential of Vision Transformers when contextual data is unavailable or omitted. The contributions of this research are as follows:
\begin{itemize}
    \item We evaluate ViT-Base on a large-scale, real-world dataset of online auto parts listings, focusing solely on visual data to isolate the model's ability to capture patterns.
    \item By clustering and analyzing image embeddings, we demonstrate ViT's effectiveness in grouping visually similar items while also identifying challenges, such as overlapping clusters and outliers.
    \item Our findings highlight the limitations of single-modal models in the absence of contextual information, offering a direct comparison to the performance of multimodal approaches.
    \item We propose directions for enhancing single-modal models and emphasize the role they can play in scenarios where multimodal inputs are unavailable or impractical.
\end{itemize}
This work offers a critical perspective on the capabilities and constraints of Vision Transformers, laying a foundation for future advancements in single-modal and hybrid approaches.

\section{Related Work}

The application of transformer architectures in computer vision has recently gained significant attention. Initially designed for natural language processing tasks, the Transformer model~\cite{vaswani2023attentionneed} demonstrated exceptional capabilities in learning relationships between sequential data, leading to its adaptation for image processing tasks. Vision Transformer~\cite{dosovitskiy2021imageworth16x16words} represents a significant shift in image classification approaches, outperforming traditional Convolutional Neural Networks (CNNs) such as ResNet~\cite{he2016deep} in various benchmarks, particularly with large-scale datasets.

The success of ViT can be attributed to its self-attention mechanism, which provides a global receptive field, unlike the local receptive fields inherent in CNNs. \citet{dosovitskiy2021imageworth16x16words} demonstrated that ViT can capture long-range dependencies across image regions, making it suitable for diverse computer vision tasks. However, ViT's dependence on large datasets for pre-training has been a known limitation, prompting studies like~\citet{touvron2021training} to introduce hybrid approaches, integrating CNNs and Transformers for improved performance on smaller datasets.

Beyond image classification, several studies have adapted the transformer architecture to other computer vision tasks. \citet{carion2020endtoendobjectdetectiontransformers} proposed DETR (DEtection TRansformer), which applies the transformer architecture to object detection, while \citet{radford2016unsupervisedrepresentationlearningdeep} leveraged transformer architectures for generative models, illustrating their versatility. Using self-attention for object detection and generative tasks further demonstrates the potential of transformer-based models in capturing intricate patterns.

Recent works have also explored multimodal approaches to tackle the challenges in computer vision tasks, particularly in understanding complex datasets like those found in online marketplaces. Multimodal models like ImageBind~\cite{girdhar2023imagebind} and OpenFlamingo~\cite{alayrac2023flamingo} have demonstrated the effectiveness of combining text and image data, which often provides richer context and improves the interpretability of results. For instance, ~\citet{hamara2024latentenginemanifoldsanalyzing} used a multimodal model for analyzing car part listings from online marketplaces, achieving higher clustering accuracy than single-modal approaches. These results highlight the limitations of single-modal models like ViT when applied to data that could benefit from context.

In the domain of combating illicit activities in online marketplaces, machine learning models have shown promise. \citet{rashid2024aisafetypracticeenhancing} utilized multimodal transformers to detect counterfeit products by integrating visual and textual cues, achieving superior performance compared to image-only models. Such studies underscore the advantage of using multimodal data for tasks requiring contextual understanding, which can be crucial in distinguishing between legitimate and illicit listings. Our work diverges from these approaches by focusing solely on the visual component, evaluating the effectiveness of ViT in detecting patterns in a single-modality context. This provides insight into the capabilities and limitations of visual-only analysis in addressing issues like the sale of stolen car parts.

Despite the promising results of multimodal approaches, single-modality models have their advantages, particularly in scenarios where only one type of data is available or where computational resources are limited. \citet{wu2020visualtransformerstokenbasedimage} argued that simplifying the input modality can reduce computational complexity and yield effective representations if the model is sufficiently trained. Thus, our study aims to contribute to this area by investigating how well a single-modality ViT model can classify and cluster car parts from online listings, identifying the strengths and areas for improvement.

Our work builds on the foundational advancements in Vision Transformers and contributes to the body of knowledge by applying ViT to a practical, real-world problem involving the analysis of car part listings. The following section discusses the ViT architecture.

\section{Overview of ViT-Base Model}

\begin{figure*}[h!]
    \centering
    \includegraphics[width=\textwidth]{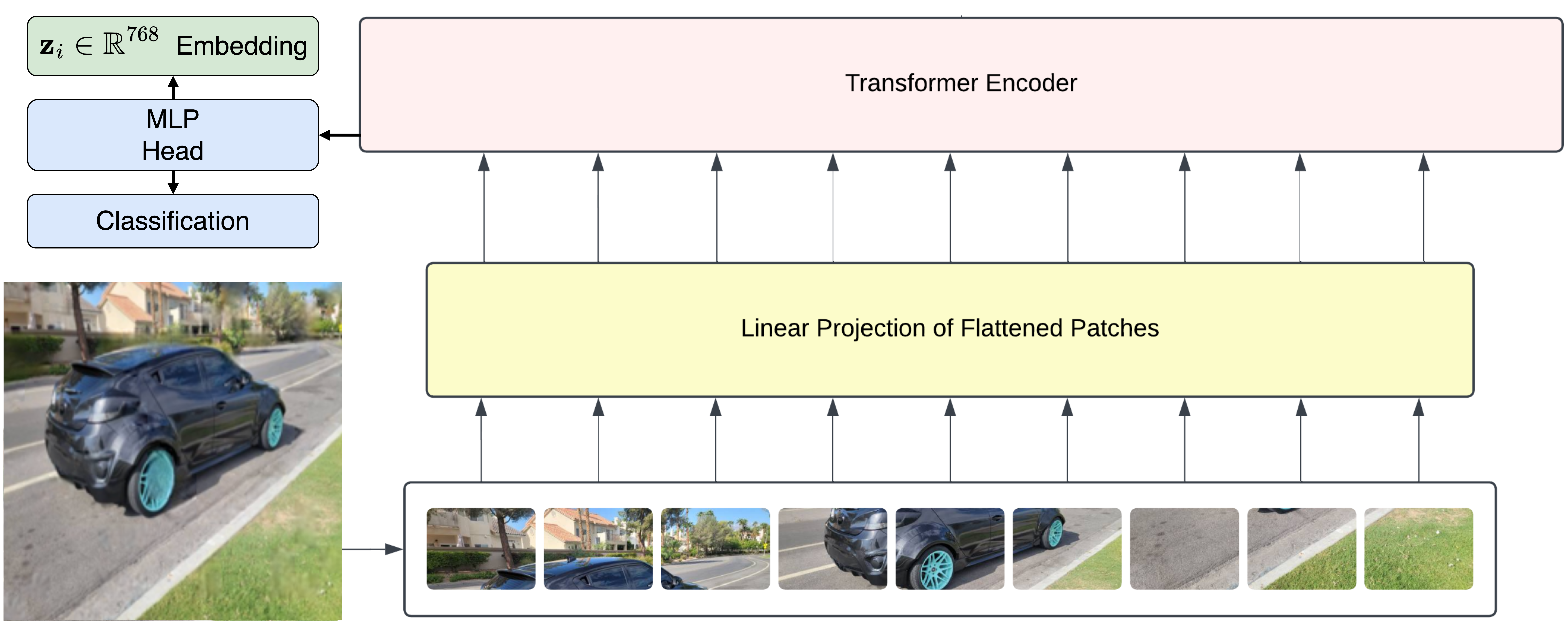}
    \caption{Vision Transformer architecture, illustrating the process of dividing an input image into patches, applying a linear projection, and processing through a transformer encoder. The final output is classified using an MLP head. Adapted from \cite{dosovitskiy2021imageworth16x16words, vaswani2023attentionneed}.}
    \label{vitdiagram}
\end{figure*}

This section provides an overview of the ViT-Base model, an implementation of the Vision Transformer architecture specifically designed for image classification. Key aspects of the model architecture, mechanisms, and training are depicted in Figure \ref{vitdiagram} and summarized next to highlight its capabilities for image classification.

\subsection{Architecture}
A ViT draws directly from the transformer architecture \cite{vaswani2023attentionneed}, adapting its self-attention mechanism for image processing tasks. Unlike CNNs, which capture spatial hierarchies through convolutional layers \cite{cordonnier2020relationshipselfattentionconvolutionallayers}, a ViT processes images as sequences of tokens. These tokens are derived from image patches and retain high-level spatial information through embeddings. 

For an input image $\mathbf{x} \in \mathbb{R}^{H \times W \times C}$, where $H$, $W$, and $C$ represent the height, width, and number of color channels respectively, the image is divided into non-overlapping patches of size $P \times P$. Each patch is flattened into a vector and linearly projected into an embedding space, forming a sequence of patch embeddings:
\[
\mathbf{z}_0 = [\mathbf{x}_{\text{class}}; \mathbf{x}_p^1 \mathbf{E}; \dots; \mathbf{x}_p^N \mathbf{E}] + \mathbf{E}_{\text{pos}},
\]
where $\mathbf{x}_{\text{class}}$ is a learnable class embedding, $\mathbf{E}$ is the patch embedding matrix, and $\mathbf{E}_{\text{pos}}$ is the positional embedding. The resulting sequence is fed into a stack of transformer layers, each consisting of multi-headed self-attention and MLP blocks, normalized with LayerNorm and connected via residual connections \cite{dosovitskiy2021imageworth16x16words}.

\subsection{Patch Embedding}
A critical innovation in ViT is its patch embedding mechanism. Instead of processing raw pixel data, ViT divides each image into patches, typically of size $16 \times 16$ pixels. Each patch $\mathbf{x}_p$ is flattened into a vector and passed through a linear projection layer:
\[
\mathbf{x}_p \mathbf{E}, \quad \mathbf{E} \in \mathbb{R}^{(P^2 \cdot C) \times D},
\]
where $D$ is the dimension of the embedding space. This transformation retains essential visual features, enabling the model to process spatial information effectively.

\subsection{Positional Encoding}
Images, unlike text, lack an inherent order. To encode spatial information, ViT introduces positional embeddings, $\mathbf{E}_{\text{pos}}$, which are added to the patch embeddings. These positional encodings can be either learnable or fixed and ensure that spatial relationships between patches are preserved, providing the transformer with context about each patch’s location within the image.

\subsection{Self-Attention}
The self-attention mechanism is central to ViT’s ability to capture global dependencies across the image. For a sequence of input embeddings $\mathbf{z} \in \mathbb{R}^{N \times D}$, self-attention computes pairwise relationships using query ($\mathbf{Q}$), key ($\mathbf{K}$), and value ($\mathbf{V}$) matrices:
\[
\text{Attention}(\mathbf{Q}, \mathbf{K}, \mathbf{V}) = \text{softmax}\left(\frac{\mathbf{Q} \mathbf{K}^\top}{\sqrt{D}}\right) \mathbf{V}.
\]
Multi-headed self-attention extends this by employing multiple attention mechanisms in parallel, allowing the model to focus on different aspects of the input. This mechanism enables ViT to capture both local and global relationships, unlike CNNs, which are limited by their receptive field \cite{cordonnier2020relationshipselfattentionconvolutionallayers}.

\subsection{Training}
Training ViT from scratch demands substantial data and computational resources due to its reliance on self-attention, which scales quadratically with the number of patches. Pre-training on large-scale datasets like ImageNet-21k provides a strong initialization, allowing fine-tuning on smaller, domain-specific datasets. In this study, ViT is fine-tuned on a car part image classification dataset, where the pre-trained classification head is replaced by a task-specific feedforward layer:
\[
\mathbf{y} = \text{softmax}(\mathbf{W} \mathbf{z}_L),
\]
where $\mathbf{z}_L$ is the output of the final transformer layer, and $\mathbf{W} \in \mathbb{R}^{D \times K}$ maps the latent space to the $K$ output classes.

By leveraging pre-training, ViT achieves robust performance even with limited task-specific data, highlighting its flexibility and effectiveness in single-modality image classification tasks.

\begin{figure*}[h!]
    \centering
    \includegraphics[width=0.8\textwidth]{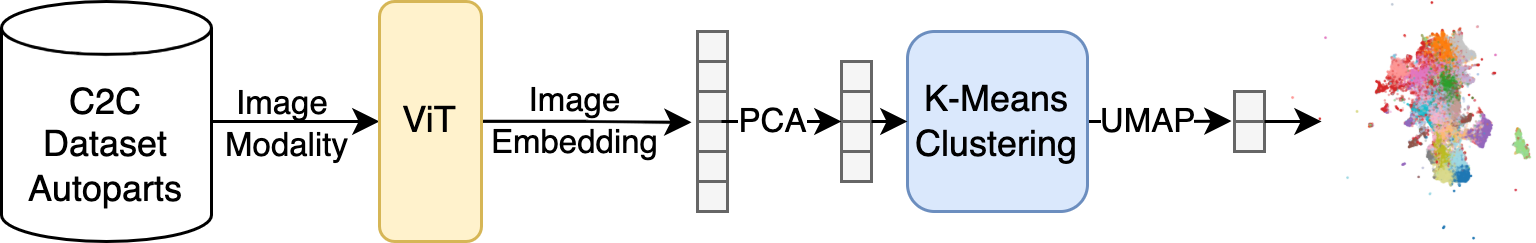}
    \caption{In the proposed methodology, the input data are images that are embedded with a ViT and then analyzed in search of cluster information.}
    \label{fig:diagram}
\end{figure*}

\section{Methodology}

This study investigates the use of Vision Transformers for clustering images of auto parts collected from online consumer-to-consumer marketplaces. The methodology depicted in Figure~\ref{fig:diagram} involves dataset acquisition, embedding extraction, clustering, and visualization. A detailed overview of each step is presented below.

\subsection{Dataset}\label{subsec:dataset}

The dataset used in this study was derived from two popular consumer-to-consumer (C2C) platforms, Craigslist and OfferUp. Posts containing car parts were identified and scraped using automated tools. From OfferUp, a total of 650,654 posts were collected, comprising approximately 500GB of images. Similarly, 637,679 posts were obtained from Craigslist, amounting to 50GB of image data. Each post often contained multiple images, providing a diverse and multimodal dataset. The complete data collection methodology, including filtering criteria, scraping procedures, and dataset structure, is detailed in our prior work \cite{hamara2024latentenginemanifoldsanalyzing}.

For this analysis, we selected a random sample of 85,000 images from the full collection, focusing exclusively on the visual component. The text data, while valuable for providing context, was intentionally excluded to evaluate the effectiveness of single-modal Vision Transformers in capturing visual patterns. This subset was processed into embeddings for clustering and visualization. 

As clustering is an unsupervised task, this study has no traditional train/val/test split. Instead, we performed a de-duplication step to ensure no duplicate images were included in the dataset, preventing potential biases in cluster formation. The dataset was used purely for exploratory analysis, with no supervised learning involved. This ensures that the clustering results reflect the inherent structure in the data rather than being influenced by a label.

The multimodal nature of the dataset, including both text and images, highlights its potential for future research involving multimodal models. However, due to privacy concerns, the dataset will remain confidential.

\subsection{Embeddings}

The pre-trained ViT-Base model, with its 12 transformer layers and 768-dimensional output space, serves as the feature extractor. For each image $\mathbf{x}_i$, the model generates a fixed-size embedding vector $\mathbf{z}_i \in \mathbb{R}^{768}$:
\[
\mathbf{z}_i = \text{ViT}(\mathbf{x}_i),
\]
where $\mathbf{z}_i$ encapsulates key visual features of $\mathbf{x}_i$. These embeddings are high-dimensional representations designed to capture semantic and structural information within the images. The embeddings were subsequently normalized and stored for downstream tasks, including dimensionality reduction and clustering.

\subsection{Clustering}

To analyze and interpret the extracted embeddings, we applied clustering methods to group similar images. Given the high dimensionality of the embeddings, we used UMAP (Uniform Manifold Approximation and Projection) for dimensionality reduction \cite{mcinnes2020umap}, projecting the data into a lower-dimensional space (2D) for visualization. UMAP preserves both local and global data structures, making it ideal for embedding analysis. 

For the clustering process, $k$-means was used due to its simplicity and effectiveness in partitioning data into non-overlapping clusters. We experimented with reduced embedding dimensions of 16, 32, 64, and 128. For each configuration, we evaluated clustering quality using three metrics:
\begin{itemize}
\item \textbf{Silhouette Score}~\cite{ROUSSEEUW198753}, which measures how similar an object is to its cluster compared to other clusters.
\item \textbf{Calinski-Harabasz Index (C-H)}~\cite{calinskiharabasz1974}, which evaluates inter-cluster variance.
\item \textbf{Davies-Bouldin Index (D-B)}~\cite{4766909}, which quantifies the average similarity between clusters.
\end{itemize}

The $k$-means objective function minimizes the within-cluster variance:
\[
J = \sum_{i=1}^{k} \sum_{\mathbf{x} \in C_i} \|\mathbf{x} - \mu_i\|^2,
\]
where $C_i$ is the set of points in cluster $i$, and $\mu_i$ is the centroid of cluster $i$. After experimentation, the optimal value of $k$ was determined to be 20, balancing intra-cluster cohesion and inter-cluster separation.

While K-Means is a widely used clustering algorithm due to its simplicity and efficiency, it has inherent limitations. Specifically, K-Means assumes that clusters are spherical and equally sized, which may not align with the underlying structure of the data. Additionally, K-Means struggles with irregularly shaped clusters and is sensitive to the initial cluster centroids. Alternative clustering methods, such as DBSCAN \cite{ester1996density} or hierarchical clustering \cite{murtagh2012algorithms}, could provide a better fit for complex structures in the data. Future work could explore these methods to assess whether they yield improved clustering quality for this dataset.

\begin{figure*}[h!]
    \centering
    \includegraphics[width=0.75\textwidth]{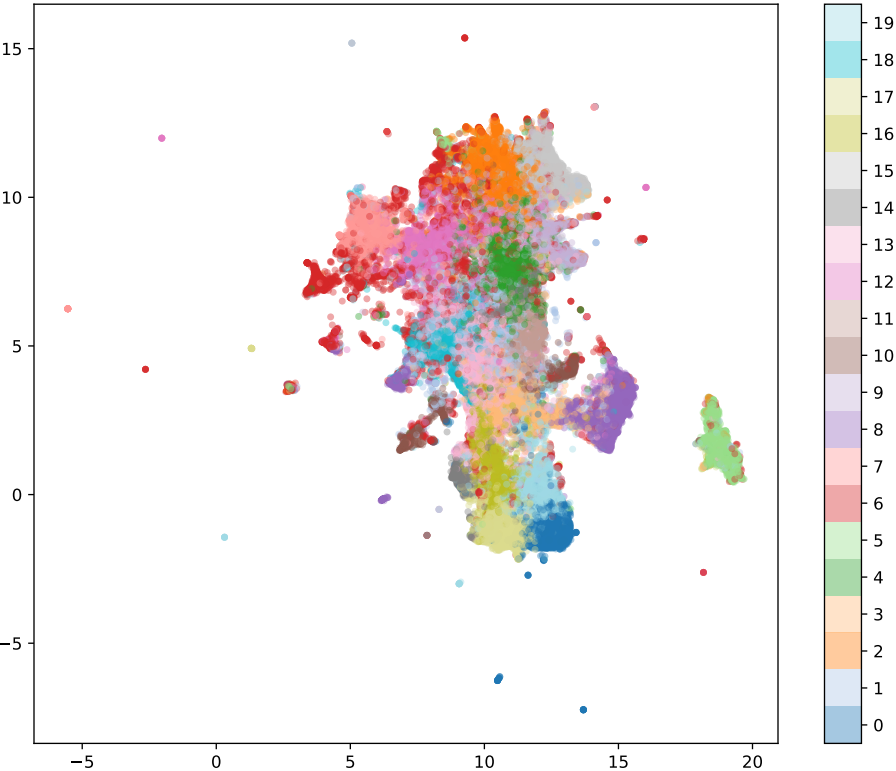}
    \caption{UMAP visualization of embeddings reduced to 64 dimensions, illustrating the clustering of images from online auto parts listings. Each color represents a distinct cluster identified using K-Means, revealing patterns and relationships within the dataset.}
    \label{umap}
\end{figure*}

\subsection{Results of Clustering Dimensions}

Our experiments revealed that using 64 dimensions provided the best overall performance. Table~\ref{tab:accents} summarizes the clustering metrics across different embedding dimensions.

\begin{table}[h!]
  \centering
  \caption{Reduced embedding dimensions along with corresponding index scores. Calinski-Harabasz and Davies-Bouldin are abbreviated as C-H and D-B, respectively.}
  \begin{tabular}{rrrr}
    \hline
    \textbf{Dim.} & \textbf{Silhouette} & \textbf{C-H} & \textbf{D-B} \\
    \hline
    16     & .0126 & 925.5 & 4.412 \\
    32     & .0134 & 937.5 & 4.274 \\
    64     & .0152 & 942.4 & 4.164 \\
    128    & .0151 & 944.3 & 4.253 \\
    \hline
  \end{tabular}
  \label{tab:accents}
\end{table}

While 128 dimensions slightly outperformed 64 in the Calinski-Harabasz Index, 64 dimensions offered the best trade-off between computational efficiency and cluster purity. The low Silhouette Scores suggest some overlap between clusters, attributed to outliers in the dataset, such as images containing mixed content (e.g., both powertrains and vehicle exteriors). This limitation highlights the challenges of single-modal approaches, especially when applied to multimodal datasets.

\subsection{Interpretation}

The clustering results indicate that ViT's embeddings can group visually similar images effectively, with a clear differentiation between major categories of auto parts. However, the presence of outliers emphasizes the importance of context, which a single-modal approach may miss. This finding underscores the potential advantages of multimodal models but also demonstrates the capability of ViT for image-only analysis in domains where additional modalities are unavailable or impractical.

The following section will analyze the findings to evaluate their implications for auto parts classification and clustering.

\section{Results and Discussion}

Our analysis revealed clusters that exhibit distinct patterns and characteristics, as shown in the UMAP visualization in Figure~\ref{umap}. These clusters highlight coherent groupings within the dataset, capturing similarities in image features while also exposing outliers that may require further investigation. The visualization demonstrates that the use of UMAP effectively reduced the high-dimensional embedding space to two dimensions, preserving important structural relationships within the data.

\subsection{Cluster Analysis}

\begin{figure}[b!]
    \centering
    \includegraphics[width=\columnwidth]{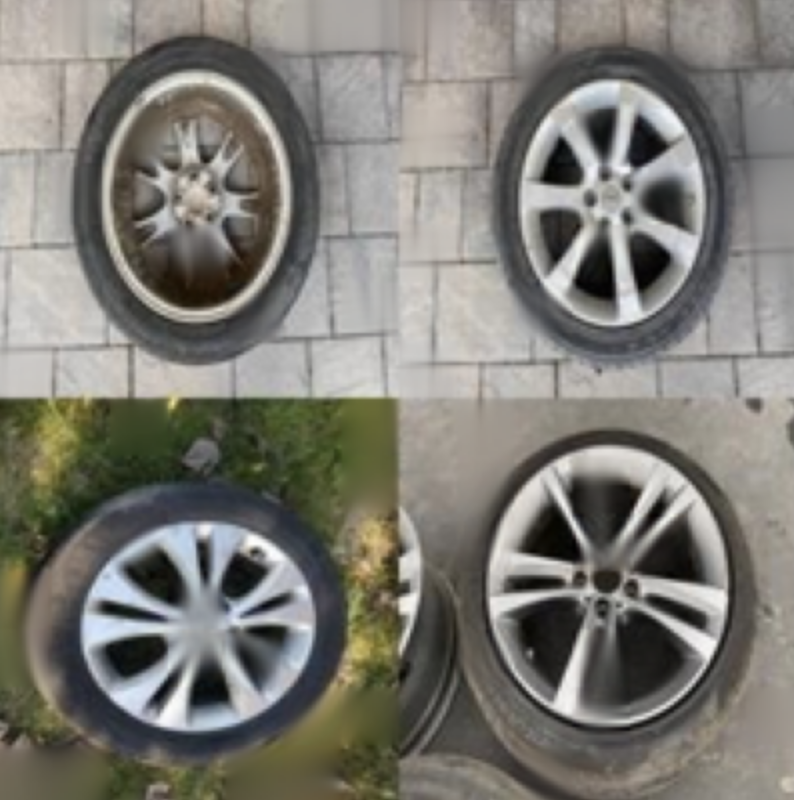}
    \caption{Representative images from posts located near a cluster centroid that appears to represent images of objects that look like \emph{wheels}.}
    \label{fig:tires}
\end{figure}

\begin{figure}[b!]
    \centering
    \includegraphics[width=\columnwidth]{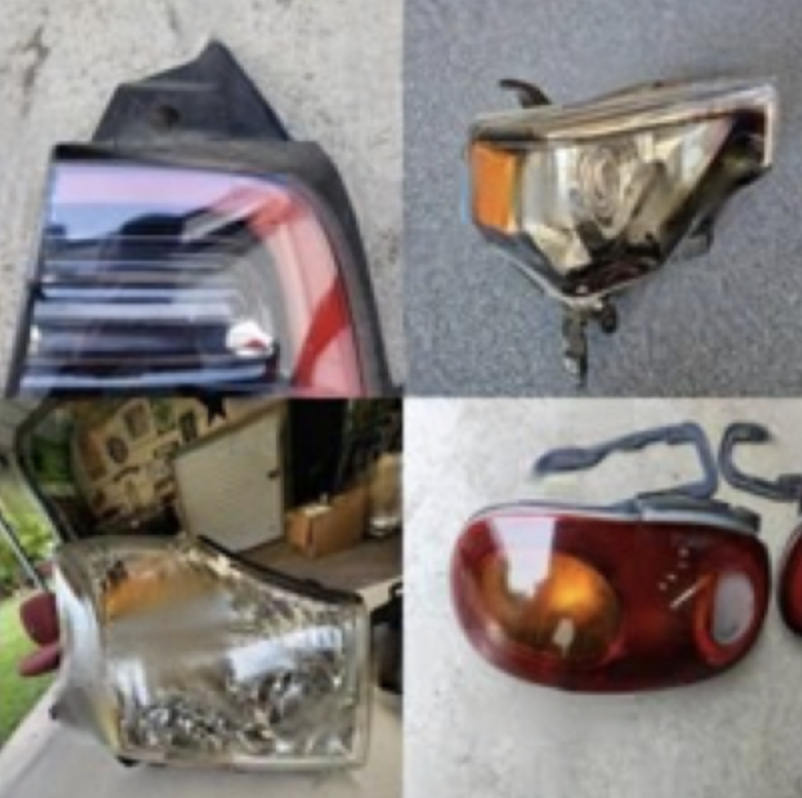}
    \caption{Representative images from posts located near a cluster centroid that appears to represent images of objects that look like \emph{lights}.}
    \label{fig:lights}
\end{figure}

\begin{figure}[b!]
    \centering
    \includegraphics[width=\columnwidth]{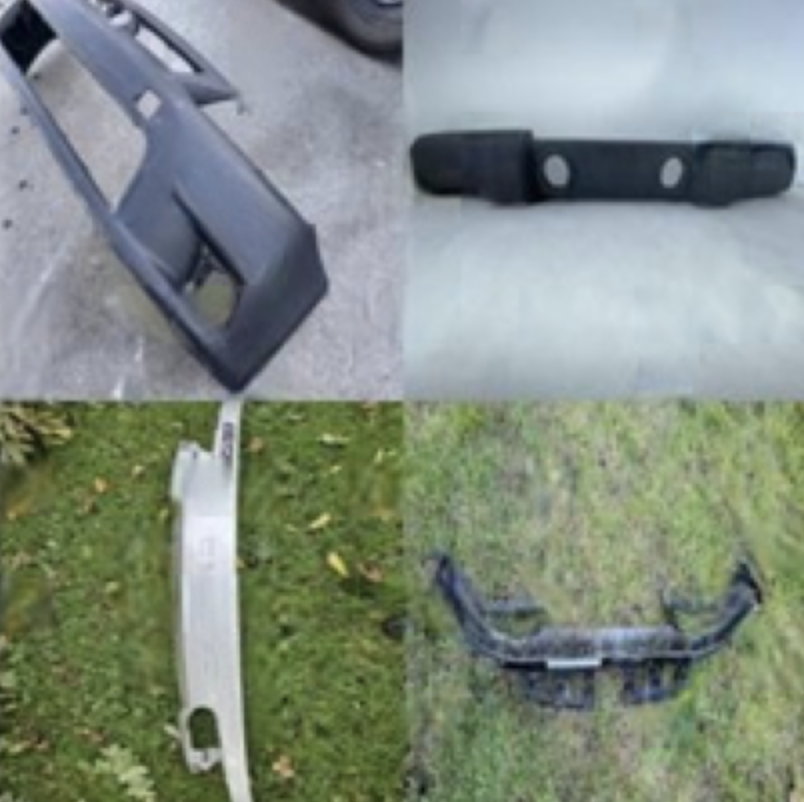}
    \caption{Representative images from posts located near a cluster centroid that appears to represent images of objects that look like \emph{bumpers}.}
    \label{fig:bumpers}
\end{figure}

Using $k$-means clustering with $k=20$, we identified distinct groups of images corresponding to specific auto part categories. To validate these clusters, we employed $k$-Nearest Neighbors (KNN), locating the ten posts nearest to each cluster centroid based on Euclidean distance. The alignment of these nearest images with their respective centroids, as displayed in Figures~\ref{fig:tires}, \ref{fig:lights}, and \ref{fig:bumpers}, supports the validity of our clustering approach.

\begin{figure}[t!]
    \centering
    \begin{subfigure}{0.48\columnwidth}
        \includegraphics[width=\columnwidth]{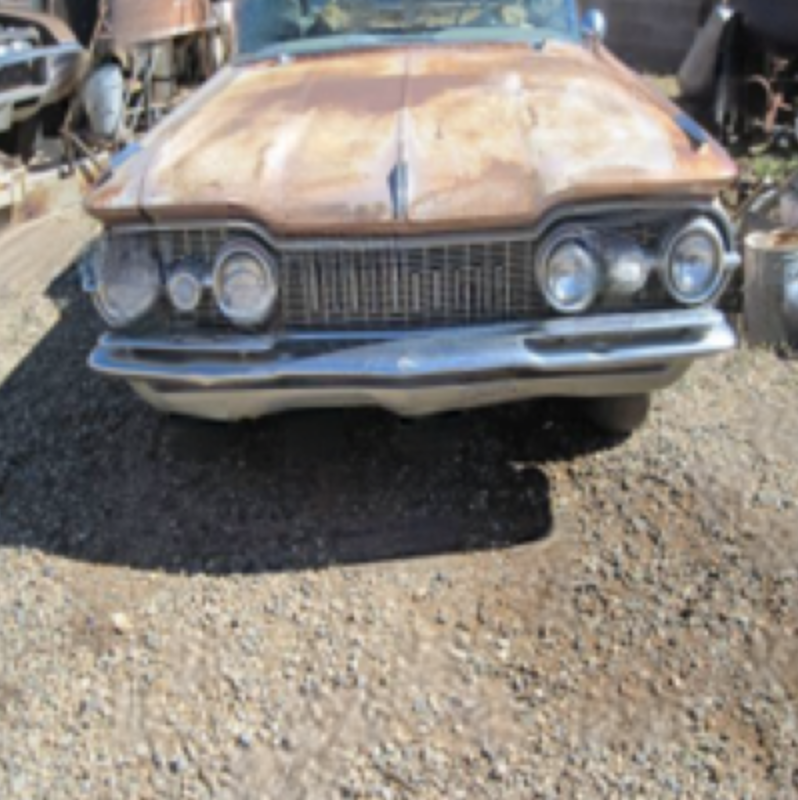}
        \caption{Hood of a vehicle, representing an image grouped in cluster~3.}
        \label{hood}
    \end{subfigure}
    \hfill
    \begin{subfigure}{0.48\columnwidth}
        \includegraphics[width=\columnwidth]{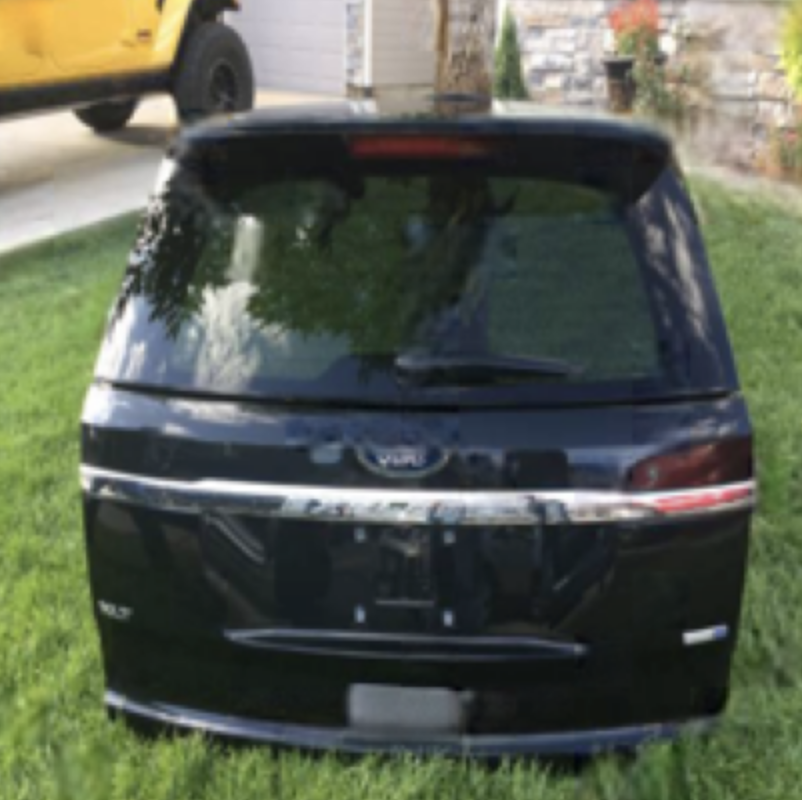}
        \caption{Trunk of a vehicle, representing another image grouped in cluster 3.}
        \label{trunk}
    \end{subfigure}
    \caption{Examples of images from cluster 3, showcasing body panels. These images highlight the visual consistency of cluster 3 in grouping related components, such as vehicle hoods and trunks.}
\end{figure}

The distinctiveness of some clusters was evident in their thematic consistency. For example:
\begin{itemize}
    \item \textbf{Cluster 0}: Primarily contained full vehicle exteriors, including sedans, SUVs, and trucks.
    \item \textbf{Cluster 1}: Dominated by individual exterior components, such as mirrors, bumpers, and grilles.
    \item \textbf{Cluster 2}: Grouped powertrain elements, including engines, transmissions, and drivetrain components.
    \item \textbf{Cluster 3}: Captured body panels like doors, trunks, and hoods, as illustrated in Figure~\ref{hood} and Figure~\ref{trunk}.
    \item \textbf{Cluster 4}: Focused on towing accessories, such as trailer hitches and tow bars.
\end{itemize}

The clustering performance metrics from Table~\ref{tab:accents} corroborated these observations. For 64 dimensions, the Calinski-Harabasz Index (942.4) indicated strong inter-cluster separation, while the Davies-Bouldin Index (4.164) suggested moderate intra-cluster cohesion. However, the low silhouette score (0.015) revealed some degree of cluster overlap, which is reflected in the presence of outliers.

The results indicate that some clusters exhibit significant overlap, which may be partially attributed to the limitations of K-Means. Since K-Means assumes spherical clusters, it may struggle to capture more complex relationships between auto-part images. For example, DBSCAN, which groups data points based on density, could potentially handle cases where clusters have irregular shapes or varying densities. Likewise, hierarchical clustering may offer a more flexible way to identify nested structures in the data. Future work should explore these methods to determine their effectiveness in improving clustering performance for online marketplace images.

Since this study employs an unsupervised clustering approach, no separate train/val/test split exists. Instead, the dataset was processed as a whole, with a de-duplication step ensuring no duplicate samples were present. This allows the clustering results to emerge naturally from the data's inherent structure rather than being influenced by training procedures. Future work could explore whether supervised approaches or hybrid methods incorporating labeled data might further refine these clusters.

\subsection{Outliers and Limitations}

The UMAP visualization also revealed a number of outliers that did not align well with any cluster. These outliers often included images with mixed or ambiguous features, such as a single image displaying both powertrain components and exterior parts. This highlights a key limitation of using a single-modal approach: while ViT is effective at capturing visual patterns, it lacks the contextual understanding that multimodal models can provide by integrating textual descriptions or other metadata.

Additionally, the low silhouette score suggests that some clusters may overlap, particularly in cases where visual similarities exist between different auto parts. For instance, mirrors and body panels may share reflective surfaces or geometric shapes that lead to misclassification. Addressing this issue would likely require incorporating additional data modalities or refining the embedding process to better differentiate these subtle features.

\subsection{Implications for Single-Modality Models}

Despite the noted limitations, the results underscore the capability of ViT to classify and group auto part images based purely on visual data. This finding is particularly relevant for applications where only image data is available or where multimodal approaches are infeasible due to computational or privacy constraints.

The ability of ViT to effectively group visually distinct categories, such as powertrain components versus exterior panels, demonstrates its potential for tasks involving large-scale, single-modality datasets. These results also highlight the utility of dimensionality reduction techniques like UMAP in improving interpretability and revealing structural relationships in high-dimensional data.

While the ViT-based approach successfully groups visually similar auto-part images, the presence of overlapping clusters and outliers suggests that additional context could improve clustering quality. A multimodal approach, such as the one explored in \cite{hamara2024latentenginemanifoldsanalyzing}, leverages both images and textual descriptions from online marketplace listings to enhance representation learning. By incorporating textual context, models like ImageBind \cite{girdhar2023imagebind} can better differentiate between visually similar objects that serve different functions (e.g., two identical-looking car parts belonging to different vehicle models). 

Future work should explore other multimodal Vision-Language Models (VLMs) that integrate textual metadata with image embeddings to refine clustering performance. This could provide richer semantic groupings and mitigate some of the cluster overlap observed in the single-modal setting.

\subsection{\emph{Big Picture} Perspective}

The findings of this study contribute to the broader understanding of single-modal architectures in computer vision. While multimodal approaches often outperform single-modal models by leveraging complementary information, this work demonstrates that a well-optimized ViT-based pipeline can achieve meaningful results in specific domains. For instance, the ability to detect patterns in C2C auto part listings could support applications in crime prevention and fraud detection.

Moreover, the challenges identified, such as cluster overlap and the handling of outliers, highlight areas for further research. Enhancing single-modal models through techniques like self-supervised learning or advanced embedding strategies could mitigate some of these issues. Future work could also explore how these methods compare to multimodal models, providing insights into the trade-offs between simplicity and performance in practical applications.

\section{Conclusions}

This analysis demonstrates that while Vision Transformers perform reasonably well in clustering online auto parts listings, the dataset poses significant challenges for a single-modal approach. Our study diverged from previous studies that employed multimodal techniques \cite{hamara2024latentenginemanifoldsanalyzing}, which leveraged the fusion of visual and textual data to capture richer contextual information. For instance, the multimodal approach achieved a silhouette score of 0.3819, significantly outperforming the 0.015 score from our single-modal ViT-based approach. This disparity underscores the advantages of multimodal models like ImageBind, which are better equipped to handle datasets containing both images and textual metadata.

Despite these limitations, our results highlight ViT's ability to isolate visual patterns in a dataset where text and images are often complementary. ViT was able to group listings into reasonably coherent clusters; however, cluster overlap and the presence of outliers revealed key shortcomings. One major limitation stems from the exclusion of textual data, such as captions, which often provide crucial contextual information about the images. Without this data, the model struggled to distinguish between visually similar components that serve different functions. Furthermore, many listings included images of multiple vehicle components within a single post, contributing to less distinct clusters and complicating the clustering process.

The findings of this study suggest several areas for improvement. One promising direction is to fine-tune the ViT model with a domain-specific pre-training dataset, allowing it to better capture the nuances of auto part imagery. Another potential enhancement involves experimenting with hybrid models that incorporate textual embeddings while maintaining a primary focus on visual data. Additionally, post-processing techniques such as outlier detection and filtering could be developed to improve the clarity and coherence of clusters.

Ultimately, the results of this study serve as a stepping stone toward developing robust tools for analyzing online marketplaces. By improving clustering performance and integrating contextual information, this approach could play a critical role in detecting patterns of illicit activity, such as identifying stolen auto parts or fraudulent listings. Continued efforts in refining single-modal and multimodal methods will help bridge the gap between theoretical advancements and practical applications in this domain.

\section{Acknowledgments}
This work was funded in part by the 
National Science Foundation 
under grants 
CNS-2210091 and CNS-2136961.

\bibliography{aaai25}

\begin{thebibliography}{21}
\providecommand{\natexlab}[1]{#1}

\bibitem[{Alayrac et~al.(2023)}]{alayrac2023flamingo}
Alayrac, J.-B.; et~al. 2023.
\newblock Flamingo: a Visual Language Model for Few-Shot Learning.
\newblock In \emph{Advances in Neural Information Processing Systems (NeurIPS)}.

\bibitem[{Caliński and Harabasz(1974)}]{calinskiharabasz1974}
Caliński, T.; and Harabasz, J. 1974.
\newblock A dendrite method for cluster analysis.
\newblock \emph{Communications in Statistics}, 3(1): 1--27.

\bibitem[{Carion et~al.(2020)Carion, Massa, Synnaeve, Usunier, Kirillov, and Zagoruyko}]{carion2020endtoendobjectdetectiontransformers}
Carion, N.; Massa, F.; Synnaeve, G.; Usunier, N.; Kirillov, A.; and Zagoruyko, S. 2020.
\newblock End-to-End Object Detection with Transformers.
\newblock arXiv:2005.12872.

\bibitem[{Cordonnier, Loukas, and Jaggi(2020)}]{cordonnier2020relationshipselfattentionconvolutionallayers}
Cordonnier, J.-B.; Loukas, A.; and Jaggi, M. 2020.
\newblock On the Relationship between Self-Attention and Convolutional Layers.
\newblock arXiv:1911.03584.

\bibitem[{Davies and Bouldin(1979)}]{4766909}
Davies, D.~L.; and Bouldin, D.~W. 1979.
\newblock A Cluster Separation Measure.
\newblock \emph{IEEE Transactions on Pattern Analysis and Machine Intelligence}, PAMI-1(2): 224--227.

\bibitem[{Dosovitskiy et~al.(2021)Dosovitskiy, Beyer, Kolesnikov, Weissenborn, Zhai, Unterthiner, Dehghani, Minderer, Heigold, Gelly, Uszkoreit, and Houlsby}]{dosovitskiy2021imageworth16x16words}
Dosovitskiy, A.; Beyer, L.; Kolesnikov, A.; Weissenborn, D.; Zhai, X.; Unterthiner, T.; Dehghani, M.; Minderer, M.; Heigold, G.; Gelly, S.; Uszkoreit, J.; and Houlsby, N. 2021.
\newblock An Image is Worth 16x16 Words: Transformers for Image Recognition at Scale.
\newblock arXiv:2010.11929.

\bibitem[{Ester et~al.(1996)Ester, Kriegel, Sander, Xu et~al.}]{ester1996density}
Ester, M.; Kriegel, H.-P.; Sander, J.; Xu, X.; et~al. 1996.
\newblock A density-based algorithm for discovering clusters in large spatial databases with noise.
\newblock In \emph{kdd}, volume~96, 226--231.

\bibitem[{Girdhar et~al.(2023)Girdhar, El-Nouby, Liu, Singh, Alwala, Joulin, and Misra}]{girdhar2023imagebind}
Girdhar, R.; El-Nouby, A.; Liu, Z.; Singh, M.; Alwala, K.~V.; Joulin, A.; and Misra, I. 2023.
\newblock Imagebind: One embedding space to bind them all.
\newblock In \emph{Proceedings of the IEEE/CVF Conference on Computer Vision and Pattern Recognition}, 15180--15190.

\bibitem[{Goodfellow et~al.(2014)Goodfellow, Pouget-Abadie, Mirza, Xu, Warde-Farley, Ozair, Courville, and Bengio}]{goodfellow2014generativeadversarialnetworks}
Goodfellow, I.~J.; Pouget-Abadie, J.; Mirza, M.; Xu, B.; Warde-Farley, D.; Ozair, S.; Courville, A.; and Bengio, Y. 2014.
\newblock Generative Adversarial Networks.
\newblock arXiv:1406.2661.

\bibitem[{Hamara and Rivas(2024)}]{hamara2024latentenginemanifoldsanalyzing}
Hamara, A.; and Rivas, P. 2024.
\newblock From Latent to Engine Manifolds: Analyzing ImageBind's Multimodal Embedding Space.
\newblock arXiv:2409.10528.

\bibitem[{He et~al.(2016)He, Zhang, Ren, and Sun}]{he2016deep}
He, K.; Zhang, X.; Ren, S.; and Sun, J. 2016.
\newblock Deep Residual Learning for Image Recognition.
\newblock In \emph{Proceedings of the IEEE conference on computer vision and pattern recognition (CVPR)}, 770--778.

\bibitem[{McInnes, Healy, and Melville(2020)}]{mcinnes2020umap}
McInnes, L.; Healy, J.; and Melville, J. 2020.
\newblock UMAP: Uniform Manifold Approximation and Projection for Dimension Reduction.
\newblock arXiv:1802.03426.

\bibitem[{Murtagh and Contreras(2012)}]{murtagh2012algorithms}
Murtagh, F.; and Contreras, P. 2012.
\newblock Algorithms for hierarchical clustering: an overview.
\newblock \emph{Wiley Interdisciplinary Reviews: Data Mining and Knowledge Discovery}, 2(1): 86--97.

\bibitem[{Radford, Metz, and Chintala(2016)}]{radford2016unsupervisedrepresentationlearningdeep}
Radford, A.; Metz, L.; and Chintala, S. 2016.
\newblock Unsupervised Representation Learning with Deep Convolutional Generative Adversarial Networks.
\newblock arXiv:1511.06434.

\bibitem[{Radford and Narasimhan(2018)}]{Radford2018ImprovingLU}
Radford, A.; and Narasimhan, K. 2018.
\newblock Improving Language Understanding by Generative Pre-Training.

\bibitem[{Rashid and Rivas(2024)}]{rashid2024aisafetypracticeenhancing}
Rashid, M.~B.; and Rivas, P. 2024.
\newblock AI Safety in Practice: Enhancing Adversarial Robustness in Multimodal Image Captioning.
\newblock arXiv:2407.21174.

\bibitem[{Rousseeuw(1987)}]{ROUSSEEUW198753}
Rousseeuw, P.~J. 1987.
\newblock Silhouettes: A graphical aid to the interpretation and validation of cluster analysis.
\newblock \emph{Journal of Computational and Applied Mathematics}, 20: 53--65.

\bibitem[{Shehzadi et~al.(2023)Shehzadi, Hashmi, Stricker, and Afzal}]{shehzadi2023objectdetectiontransformersreview}
Shehzadi, T.; Hashmi, K.~A.; Stricker, D.; and Afzal, M.~Z. 2023.
\newblock Object Detection with Transformers: A Review.
\newblock arXiv:2306.04670.

\bibitem[{Touvron et~al.(2021)Touvron, Cord, Douze, Massa, Sablayrolles, and J{\'e}gou}]{touvron2021training}
Touvron, H.; Cord, M.; Douze, M.; Massa, F.; Sablayrolles, A.; and J{\'e}gou, H. 2021.
\newblock Training data-efficient image transformers \& distillation through attention.
\newblock In \emph{International Conference on Machine Learning (ICML)}, 10347--10357.

\bibitem[{Vaswani et~al.(2023)Vaswani, Shazeer, Parmar, Uszkoreit, Jones, Gomez, Kaiser, and Polosukhin}]{vaswani2023attentionneed}
Vaswani, A.; Shazeer, N.; Parmar, N.; Uszkoreit, J.; Jones, L.; Gomez, A.~N.; Kaiser, L.; and Polosukhin, I. 2023.
\newblock Attention Is All You Need.
\newblock arXiv:1706.03762.

\bibitem[{Wu et~al.(2020)Wu, Xu, Dai, Wan, Zhang, Yan, Tomizuka, Gonzalez, Keutzer, and Vajda}]{wu2020visualtransformerstokenbasedimage}
Wu, B.; Xu, C.; Dai, X.; Wan, A.; Zhang, P.; Yan, Z.; Tomizuka, M.; Gonzalez, J.; Keutzer, K.; and Vajda, P. 2020.
\newblock Visual Transformers: Token-based Image Representation and Processing for Computer Vision.
\newblock arXiv:2006.03677.

\end{thebibliography}

\end{document}